\documentclass[]{fairmeta}

\usepackage{amsmath,amssymb,amsfonts}
\usepackage{fvextra}
\usepackage{tabularx}
\usepackage{algorithm}
\usepackage{algpseudocode}
\usepackage{enumitem}
\usepackage{needspace}
\usepackage{flafter}
\setlist[itemize]{leftmargin=*}
\DeclareUnicodeCharacter{2010}{-}
\DeclareUnicodeCharacter{2011}{-}
\DeclareUnicodeCharacter{2013}{--}
\DeclareUnicodeCharacter{2014}{---}
\DeclareUnicodeCharacter{2019}{'}
\DeclareUnicodeCharacter{201C}{``}
\DeclareUnicodeCharacter{201D}{''}
\DeclareUnicodeCharacter{2026}{\ldots}

\newcommand{\hlbox}[2]{%
  \begingroup%
  \setlength{\fboxsep}{1.5pt}%
  \colorbox{#1}{\vphantom{Ay}#2}%
  \endgroup%
}
\newcommand{\stoken}[1]{\texttt{\textless#1\textgreater}}

\definecolor{usergray}{RGB}{192, 192, 192}
\definecolor{tablue}{RGB}{227, 253, 253}
\definecolor{finalblue}{RGB}{113, 201, 206}

\newtcolorbox{promptbox}[1][]{%
  enhanced, breakable,
  colback=black!2, colframe=black!12, boxrule=0.4pt,
  left=6pt,right=6pt,top=6pt,bottom=6pt,
  #1
}

\title{Spoken Language Models that Think Aloud}

\author[1,2,*]{Junyi Ao}
\author[1]{Kainan Peng}
\author[1]{Mingbo Ma}
\author[1]{Shun Zhang}
\author[1]{Zhenyu Tang}
\author[1]{Xutai Ma}
\author[1]{Xiang Li}
\author[1]{Yinghao Li}
\author[1,2,*]{Yuancheng Wang}
\author[2]{Zhizheng Wu}
\author[2]{Haizhou Li}
\author[1]{Qing He}
\author[1]{Xubo Liu}

\affiliation[1]{Meta Superintelligence Labs}
\affiliation[2]{The Chinese University of Hong Kong, Shenzhen}
\contribution[*]{Work done at Meta}

\abstract{
While Chain-of-Thought (CoT) reasoning has improved the capability of language models, directly applying it to Spoken Language Models (SLMs) may introduce long silent intervals under the serial ``think-then-speak'' paradigm, disrupting real-time spoken interaction.
To address this issue, we propose an asynchronous think-aloud framework for reasoning-based SLMs within the Thinker-Talker architecture.
The framework maintains a primary reasoning stream for logical deduction and a lightweight think-aloud stream that generates short, task-grounded progress utterances conditioned on the user input and the evolving reasoning state. A dynamic balance strategy coordinates the two streams at runtime, triggering additional think-aloud speech to avoid silent gaps and canceling pending utterances when the final response becomes ready. Experiments on spoken reasoning and question-answering benchmarks show that our approach substantially reduces user-audible silence during reasoning while maintaining answer accuracy comparable to that of a serial ``think-then-speak'' baseline, demonstrating the potential of asynchronous think-aloud for responsive interaction in SLMs.}

\date{\today}
\correspondence{
 \email{junyiao1@link.cuhk.edu.cn, xuboliu@meta.com}
}
\metadata[Keywords]{spoken language model, chain of thought, human-computer interaction}

\begin{document}

\maketitle

\section{Introduction}
\label{section:intro}
\begin{figure}[!t]
     \centering
     \includegraphics[width=0.9\linewidth]{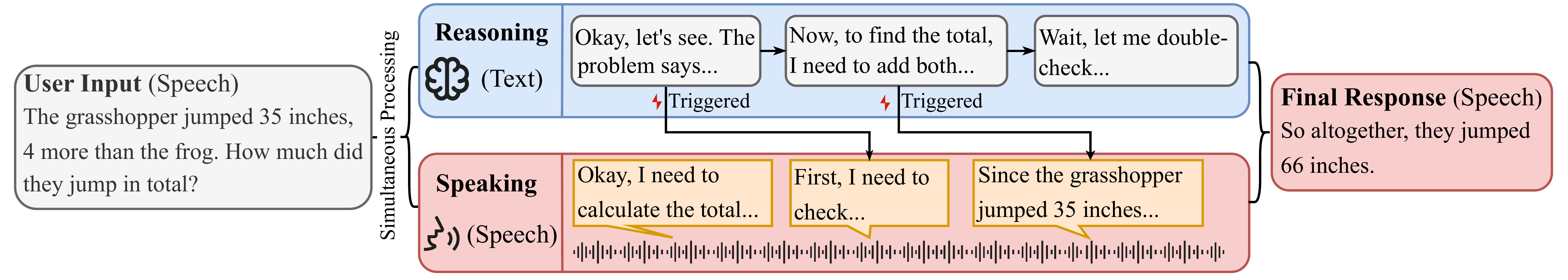}
     \caption{Illustration of the asynchronous reasoning and think-aloud generation mechanism. In this example, three think-aloud responses are generated: the first is forcibly triggered at the beginning of reasoning to bridge the start-up gap, and two additional responses are triggered at intermediate reasoning milestones.}
     \label{fig:intro}
\end{figure}
Human speakers rarely plan an entire response silently before delivering it \citep{levelt1993speaking}.
Instead, they often formulate, revise, and refine ideas within the flow of conversation.
Disfluencies, such as pauses, hedges, and self-corrections, facilitate turn-taking and signal confidence, progress, and intent in real time \citep{CLARK200273}.
Speakers also organize content incrementally, often giving a coarse structure before filling in details.
This observation motivates spoken systems that can provide timely feedback while their underlying reasoning is still unfolding.

Complex reasoning has become a cornerstone of recent LLM research \citep{qwq32b,guo2025deepseek,jaech2024openai,wei2022chain}, driving strong performance improvements in mathematical problem-solving and code generation \citep{guo2025deepseek,deepseek-math-v2,jaech2024openai}.
As spoken language models (SLMs) are increasingly expected to support question answering, tutoring, task-oriented assistance, and other interactive applications, incorporating such reasoning capabilities into speech-based interaction becomes an important next step.

However, reasoning-based spoken interaction introduces a latency-accuracy tension.
A straightforward approach is to perform full textual reasoning before speaking and then generate the final spoken response.
While this serial ``think-then-speak'' paradigm can preserve the benefits of explicit reasoning, it often produces extended silent intervals that make the system appear unresponsive.
We refer to such user-audible dead air as unmasked silence, i.e., intervals during which internal reasoning is still ongoing but no speech is available to the user.
Such silence can last many seconds for complex queries, disrupting turn-taking and weakening conversational synchrony.

This work focuses on spoken agents that require non-trivial internal reasoning, such as multi-step reasoning, knowledge-intensive question answering, tutoring, and task-oriented assistance.
For short conversational exchanges, explicitly verbalizing intermediate reasoning may be unnecessary or even undesirable.
Our goal is to improve interaction in scenarios where the model requires additional reasoning time.
In such cases, short and task-grounded progress feedback can be preferable to prolonged silence.
Therefore, we study how to reduce unmasked silence during spoken reasoning and knowledge-intensive question answering while preserving the answer quality of a serial ``think-then-speak'' baseline without exposing raw reasoning traces.

We borrow the term ``thinking aloud'' from cognitive psychology and protocol analysis \citep{ericsson1984protocol,newell1972human}, but use it for a different research purpose.
In this work, think-aloud responses are short intermediate utterances generated during the model's reasoning phase to provide audible progress feedback.
They are not intended to expose the full chain of thought.
Instead, they provide concise signals such as a task reformulation, an intermediate reasoning milestone, or a transition cue, while the main reasoning process continues in the background.

As illustrated in Figure \ref{fig:intro}, our paradigm initiates simultaneous processing upon receiving user input.
A reasoning stream performs textual deduction, while a speaking stream produces selected think-aloud utterances conditioned on the user input and, when available, the evolving reasoning state.
The system first generates an initial utterance to bridge the start-up gap, retaining the key benefit of generic fillers such as ``Let me think'': providing immediate audible feedback before the final answer is ready.
Different from a fixed filler, however, this utterance is grounded in the user's query and acknowledges the specific task being addressed.

As reasoning progresses, intermediate milestones can trigger additional speech segments.
These later utterances further provide lightweight, task-grounded progress cues while continuing to mask long silent intervals.
Unlike fixed chunk-level interleaving, this paradigm decouples reasoning progression from speech realization.
The reasoning stream can continue advancing the main reasoning trajectory, while the speaking stream verbalizes only selected progress signals.

To instantiate this paradigm, we introduce a framework based on the Thinker-Talker architecture.
The framework includes a lightweight think-aloud module that coordinates with the main reasoning thinker.
The reasoning thinker generates the internal reasoning trajectory and emits think-aloud trigger tokens at selected reasoning milestones.
The think-aloud module then produces concise intermediate utterances, which are synthesized by a unified talker.
During inference, a dynamic balance strategy decides when additional think-aloud speech should be generated or canceled.

In summary, our main contributions are as follows:
\begin{itemize}
    \item We introduce an asynchronous think-aloud framework that provides timely spoken feedback during reasoning, starting with a brief acknowledgement of the user's request and followed by concise, reasoning-grounded progress updates at selected milestones, without verbalizing the full reasoning trace.

    \item We introduce a lightweight think-aloud module within the Thinker-Talker architecture and design a dynamic balance strategy to coordinate reasoning and speaking at runtime.
    The strategy triggers additional think-aloud speech when silence emerges and cancels pending utterances when the final response becomes ready, reducing both unmasked silence and unnecessary articulation computational overhead after the reasoning completes.

    \item Experiments on spoken reasoning and question-answering benchmarks show that our approach substantially reduces unmasked silence while maintaining performance comparable to a serial ``think-then-speak'' baseline across both evaluation categories.
    These results support more responsive spoken interaction.
\end{itemize}

\begin{figure}[!t]
     \centering
     \includegraphics[width=0.9\linewidth]{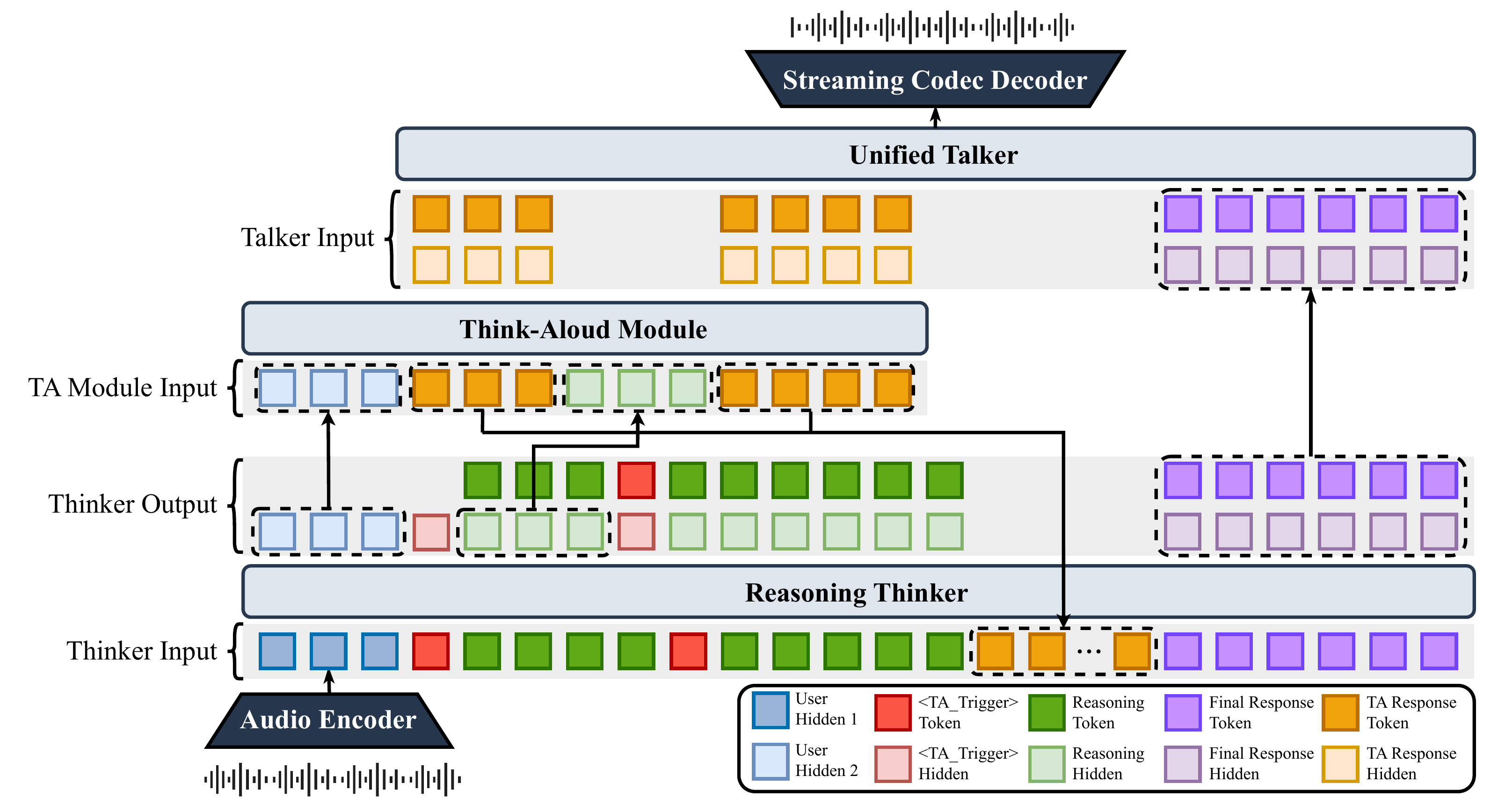}
     \caption{The overall architecture of our proposed model, illustrating the collaborative workflow between the reasoning thinker, the think-aloud module, and the unified talker. ``$\rightarrow$'' denotes the copy operation. The dashed box indicates the source and target of the copy mechanism.}
     \label{fig:overall}
\end{figure}

\section{Method}
We propose a unified architecture that separates internal reasoning from audible progress-feedback generation while coordinating them asynchronously in reasoning-based SLMs.
As shown in Figure \ref{fig:overall}, the architecture comprises three jointly optimized components: the reasoning thinker, the think-aloud module, and the unified talker.
We further propose a dynamic balance algorithm to align reasoning latency with think-aloud response duration during inference.

\subsection{The Reasoning Thinker}
The reasoning thinker serves as the cognitive core of our system.
It accepts the system prompt and the audio representations labeled as ``User Hidden 1'' in Figure \ref{fig:overall} from the audio encoder as input.
Similar to text-based LLMs with reasoning capabilities \citep{guo2025deepseek,wei2022chain}, the reasoning thinker first generates a reasoning trajectory.
Subsequently, prior to generating the final response, the think-aloud responses produced by the think-aloud module are appended to the reasoning context.
This ensures that the final response remains consistent with the think-aloud content, facilitating a more coherent generation process.

Let $\mathbf{H}^{user}$ denote the sequence of hidden state representations corresponding to the user input, which is labeled as ``User Hidden 2'' in Figure~\ref{fig:overall}.
To address the latency associated with reasoning generation, we employ a special token, \stoken{TA\_trigger}, to activate the think-aloud module.
Immediately upon processing $\mathbf{H}^{user}$, the reasoning thinker generates the first \stoken{TA\_trigger}, prompting the think-aloud module to produce an initial response before the reasoning process commences.
As the thinker proceeds to generate the reasoning trajectory, it predicts additional \stoken{TA\_trigger} tokens at specific positions, allowing the think-aloud module to update its response based on the evolving reasoning context.
Crucially, the reasoning thinker and the think-aloud module operate asynchronously; the thinker continues to generate subsequent reasoning tokens in the background, running in parallel with the think-aloud module.

\subsection{The Think-Aloud Module}
The think-aloud module is a 0.5B lightweight LLM designed to generate responses aligned with user input and reasoning content.
It processes context progressively, keeping the generated response coherent with the user query, previous think-aloud responses, and the reasoning process.
Because the reasoning thinker and the think-aloud module operate in different semantic spaces with distinct dimensions, we introduce a projection layer, $\phi_{in}: \mathbb{R}^{d_{thinker}} \to \mathbb{R}^{d_{TA}}$.

Let $k$ denote the index of the think-aloud response.
The input $\mathbf{X}^{TA}_k$ for the token sequence of the $k$-th response is constructed as follows:
\begin{itemize}
    \item \textbf{For the 1st think-aloud response ($k=1$)}:
This response is triggered immediately after the user input. Thus the input consists solely of the projected representation of the user input:$$\mathbf{X}^{TA}_1 = \phi_{in}(\mathbf{H}^{user})$$
    \item \textbf{For subsequent think-aloud responses ($k > 1$)}:
These responses are triggered upon the completion of specific reasoning segments.
The input aggregates the original user input, the history of prior think-aloud utterances, and the cumulative reasoning hidden states:
$$\mathbf{X}^{TA}_k = \left[ \mathbf{X}^{TA}_{k-1} ; Y^{TA}_{k-1} ; \phi_{in}(\mathbf{H}^{r}_{k-1} )\right]$$
where $\mathbf{H}^{r}_i$ (``Reasoning Hidden'' in Figure \ref{fig:overall}) denotes the reasoning segment generated between the $i$-th and $(i+1)$-th \stoken{TA\_trigger} token, and $\mathbf{Y}^{TA}_{i}$ (``TA Response Token'' in Figure \ref{fig:overall}) denotes the token sequence of the $i$-th think-aloud response.
\end{itemize}

Before the talker generates the final response, the think-aloud response generated by the think-aloud module is appended to the reasoning content, ensuring that the final response generated by the thinker is consistent with the user's question, the reasoning content, and the think-aloud responses.

Generic fillers such as ``Let me think'' are a simple and effective way to mask the initial silence in spoken interaction.
Our framework retains this advantage through the first forced think-aloud trigger, which is generated immediately after the user query and functions as a brief acknowledgement of the user query rather than a fixed template filler.
This first utterance plays a filler-like role: it provides fast audible feedback before the main reasoning trajectory is completed.

Later think-aloud utterances extend this mechanism beyond initial silence masking. Instead of repeatedly producing generic fillers, they are conditioned on the evolving reasoning state and triggered at selected reasoning milestones.
In this sense, the proposed framework can be interpreted as a hybrid feedback policy: it combines the low-latency silence-masking role of fillers with reasoning-grounded progress feedback for longer reasoning turns.

\subsection{The Unified Talker}
To synthesize continuous speech from the outputs of both the reasoning thinker and the think-aloud module, we employ CosyVoice 2.0 \citep{du2024cosyvoice} in streaming mode as a unified talker. To integrate linguistic information (text embeddings) with high-level guidance (hidden states), we introduce two distinct linear projection heads that map the hidden states into the talker's input dimension ($d_{talk}$). Specifically, $\phi_{out}^{th}(\cdot)$ maps $\mathbb{R}^{d_{thinker}} \to \mathbb{R}^{d_{talk}}$, and $\phi_{out}^{TA}(\cdot)$ maps $\mathbb{R}^{d_{TA}} \to \mathbb{R}^{d_{talk}}$.

At each generation step, the input $\mathbf{u}$ to the unified talker is constructed by summing the native CosyVoice text embedding $\mathbf{e}_{text}$ and the aligned representation $\mathbf{r}_r$. The source of this representation switches dynamically based on the current generation phase:
\[
\begin{aligned}
\mathbf{u} &= \mathbf{e}_{text} + \mathbf{r}_r, \\
\mathbf{r}_r &=
\begin{cases}
\phi^{TA}_{out}(\mathbf{H}^{TA}), & \text{in the reasoning phase},\\
\phi^{th}_{out}(\mathbf{H}^{th}), & \text{in the final response phase}.
\end{cases}
\end{aligned}
\]

During the reasoning phase, the talker synthesizes the think-aloud response driven by the think-aloud module's hidden states $\mathbf{H}^{TA}$ (``TA Response Hidden'' in Figure \ref{fig:overall}) to bridge silence.
Once reasoning is complete, the model enters the final response phase, where the talker uses the thinker's hidden states $\mathbf{H}^{th}$ (``Final Response Hidden'' in Figure \ref{fig:overall}) to deliver the answer.

To ensure alignment and compatibility with CosyVoice 2.0, we follow its original streaming data formatting protocol.
The inputs and outputs for the talker are prepared as interleaved sequences of input representations and output tokens at a ratio of 5 to 15, which is kept unchanged from the initialized CosyVoice 2.0 streaming configuration to avoid introducing an additional architectural variable.

\subsection{Training Objective}
To jointly optimize the reasoning thinker, the think-aloud module, and the unified talker, we employ a multi-task training objective composed of three terms, corresponding to the thinker generation, think-aloud generation, and talker generation, respectively:
$$
\mathcal{L} = \mathcal{L}_{Thinker} + \mathcal{L}_{TA} + \mathcal{L}_{Talk}
$$
Note that the audio encoder is fixed during training.

\subsection{Dynamic Balance Strategy}
\label{sec:dynamic_balance}
During training, the model learns to trigger appropriate think-aloud responses based on reasoning milestones, as described in Section~\ref{data_prep}. However, during real-time streaming inference, the physical duration of synthesized speech may not align with the varying computation time required for reasoning generation. To reduce long silent gaps and excessive articulation computational overhead, we introduce a state-driven dynamic balance strategy to coordinate the asynchronous reasoning (Thinker) and speaking (think-aloud module) streams.

Rather than relying on pre-computed time estimates, the proposed strategy tracks the runtime states of both streams and is invoked at key transition events, e.g., when the currently active think-aloud playback finishes while reasoning is still ongoing, or when the reasoning stream reaches completion. It handles two primary boundary scenarios:

\begin{itemize}
\item \textbf{Audio starvation (reasoning is ongoing, but speech playback has finished):}
This occurs when the reasoning process is still generating tokens, but no think-aloud audio remains for playback.
To reduce user-perceived silence, the strategy scans the current reasoning buffer for the latest completed logical segment that has not yet triggered a response, using the same double-newline-based segment boundary convention as in training.
It then issues a new think-aloud trigger for that segment, providing additional audible feedback while reasoning continues.

\item \textbf{Reasoning early completion (reasoning finishes while think-aloud is ongoing):}
This occurs when the thinker completes its deduction before all potential think-aloud responses have been synthesized.
To produce the final response, the strategy immediately cancels all \textit{pending} (i.e., not yet synthesized) \texttt{<TA\_trigger>} tokens.
\textbf{The currently active think-aloud utterance, however, is allowed to finish before final-response generation begins.}
The completed utterance is appended to the reasoning context so that the subsequent final response remains consistent with the speech already presented to the user.

\end{itemize}

\begin{algorithm}[!t]
\small
\caption{Dynamic Balance Strategy for Asynchronous Think-Aloud Generation}
\label{alg:dynamic_balance}
\begin{algorithmic}[1]
\Require Reasoning thinker $\mathcal{M}_{\mathrm{th}}$, think-aloud module $\mathcal{M}_{\mathrm{TA}}$, unified talker $\mathcal{T}$, user input representation $H^{\mathrm{user}}$
\Ensure Final spoken response

\State Initialize reasoning token buffer $\mathcal{R}_{r} \gets \emptyset$
\State Initialize reasoning representation buffer $\mathcal{H}_{r} \gets \emptyset$
\State Initialize completed think-aloud set $\mathcal{C}_{\mathrm{TA}} \gets \emptyset$
\State Initialize candidate prefix set $\mathcal{P}_{\mathrm{cand}} \gets \emptyset$
\State Initialize executed prefix set $\mathcal{P}_{\mathrm{exec}} \gets \emptyset$
\State $\mathrm{reasoning\_done} \gets \mathrm{False}$
\State $\mathrm{ready\_for\_final} \gets \mathrm{False}$

\State Start $\mathcal{M}_{\mathrm{th}}$ to generate the reasoning trajectory
\State Force the first \texttt{<TA\_trigger>} after receiving $H^{\mathrm{user}}$
\State Generate the first think-aloud utterance conditioned on $H^{\mathrm{user}}$
\State Start playback with the unified talker $\mathcal{T}$

\While{not $\mathrm{ready\_for\_final}$}
    \State Append newly generated reasoning tokens to $\mathcal{R}_{r}$
    \State Append their corresponding hidden representations to $\mathcal{H}_{r}$

    \If{$\mathcal{M}_{\mathrm{th}}$ emits a \texttt{<TA\_trigger>} after reasoning prefix boundary $p$}
        \State $\mathcal{P}_{\mathrm{cand}} \gets \mathcal{P}_{\mathrm{cand}} \cup \{p\}$
    \EndIf

    \If{$\mathcal{M}_{\mathrm{th}}$ finishes reasoning}
        \State $\mathrm{reasoning\_done} \gets \mathrm{True}$
        \State Cancel all pending, not-yet-synthesized \texttt{<TA\_trigger>} tokens
        \State Discard all pending candidate prefixes in $\mathcal{P}_{\mathrm{cand}} \setminus \mathcal{P}_{\mathrm{exec}}$
        \If{no think-aloud utterance is currently being played}
            \State $\mathrm{ready\_for\_final} \gets \mathrm{True}$
        \EndIf
    \EndIf

    \If{current think-aloud playback finishes}
        \State Append the completed think-aloud utterance to $\mathcal{C}_{\mathrm{TA}}$

        \If{$\mathrm{reasoning\_done}$} \Comment{Reasoning early completion}
            \State $\mathrm{ready\_for\_final} \gets \mathrm{True}$
        \Else \Comment{Audio starvation}
            \State $p^\star \gets$ latest prefix in $\mathcal{P}_{\mathrm{cand}} \setminus \mathcal{P}_{\mathrm{exec}}$

            \If{$p^\star$ does not exist}
                \State $p^\star \gets$ latest completed reasoning prefix boundary in $\mathcal{H}_{r}$ not in $\mathcal{P}_{\mathrm{exec}}$
            \EndIf

            \If{$p^\star$ exists}
                \State $\mathcal{P}_{\mathrm{exec}} \gets \mathcal{P}_{\mathrm{exec}} \cup \{p^\star\}$
                \State Generate a new think-aloud utterance conditioned on $H^{\mathrm{user}}$, $\mathcal{C}_{\mathrm{TA}}$, and $\mathcal{H}_{r}^{\le p^\star}$
                \State Start playback with the unified talker $\mathcal{T}$
            \EndIf
        \EndIf
    \EndIf
\EndWhile

\State Generate the final response conditioned on $\mathcal{R}_{r}$ and $\mathcal{C}_{\mathrm{TA}}$
\State Speak the final response with the unified talker $\mathcal{T}$

\end{algorithmic}
\end{algorithm}

The empirical effect of this strategy is analyzed in Section \ref{sec:latency}.
The latency results in Figure \ref{fig:latency} show that the resulting unmasked silence remains very low for most test samples.
Algorithm~\ref{alg:dynamic_balance} provides the pseudocode, and Section~\ref{sec:case_study} illustrates its behavior.

\section{Data Preparation}
\label{data_prep}
In this section, we describe the methodology for preparing our reasoning and ``think-aloud'' datasets using in-house SFT data, which contains both single-turn and multi-turn spoken dialogue data.

\subsection{Reasoning Data Generation}
Our reasoning data is generated through a three-step pipeline using DeepSeek-R1 \citep{guo2025deepseek} for all generation and filtering.
\begin{itemize}
    \item We first identify and select reasoning-intensive turns from the in-house SFT data.
    We evaluate each turn five times using the identical prompt, and the result is determined by a majority vote.
    \item Second, we prompt DeepSeek-R1 with the dialogue history and the current user input for each selected turn.
    We then extract the resulting reasoning text while discarding the model's final reply.
    \item Finally, we validate the generated reasoning text.
    Since the extracted reasoning text is conditioned only on the dialogue history and user input, its alignment with the final ground-truth answer is not guaranteed.
    Therefore, we apply an additional filtering pass to ensure this consistency and filter out any mismatches.
    Similar to the first step, we run five times using the same prompt and decide the answer by a majority vote.
\end{itemize}

The prompts for all three steps are provided in Appendix~\ref{sec:reasoning_prompts}.

\subsection{Think-Aloud Response Generation}
\label{ta_data}
For the generation of think-aloud responses, we employ a process with four steps.
All steps use the DeepSeek-R1 model \citep{guo2025deepseek}.
\begin{itemize}
    \item First, the reasoning text is segmented into multiple paragraphs using double newline characters as delimiters.
    Following segmentation, we identify potential trigger points for the think-aloud responses.
    Our criterion is whether a preliminary conclusion can be drawn from the reasoning content of the current segment.
    To ensure the stability of this process and filter out spurious triggers, the identification routine is executed five times, and the final set of triggers is determined via a majority vote.
    \item Once the triggers are finalized, the model generates the think-aloud responses.
    This generation is conditioned on both the original user input and the specific reasoning content associated with the validated trigger points.
    We intentionally restrict each think-aloud response to a single sentence to keep the verbalized reasoning concise, controllable in duration, and less likely to dominate the final answer during response generation.
    \item In the third step, we prompt DeepSeek-R1 to minimally rewrite the original final response conditioned on the generated think-aloud utterances, ensuring consistency with the audible progress feedback while preserving the original answer semantics.
    \item Finally, we utilize the internal TTS to resynthesize speech for all responses, ensuring acoustic consistency across the dataset.
    This process covers the entire dialogue history, the think-aloud response, and the final responses.
    All generated speech segments are unified into a single-speaker voice.
\end{itemize}

The prompts for trigger identification, utterance generation, and final-response rewriting are provided in Appendix~\ref{sec:ta_prompts}.

\FloatBarrier
\section{Experimental Setup}
\subsection{Implementation Details}
\subsubsection{Data}
We utilize a large-scale proprietary in-house dataset comprising approximately 200,000 single- and multi-turn dialogues, totaling around 5,000 hours of speech.
Due to data governance and privacy restrictions, the training corpus cannot be publicly released.
To provide context on the data distribution, these dialogues span diverse scenarios, including general commonsense QA, reasoning and logical deduction, as well as general helpfulness queries.
This composition ensures the model's coverage across diverse spoken-interaction scenarios.
The reasoning text and think-aloud responses are prepared following the methodology outlined in Section \ref{data_prep}.
To improve transparency, we provide the data-construction prompts in Appendix~\ref{sec:prompts}, the evaluation protocols in Section~\ref{sec:evaluation}, and representative outputs in Section~\ref{sec:case_study}.

\subsubsection{Model Training}
For model initialization, we leverage several pre-trained models.
The audio encoder and reasoning thinker are initialized from the Qwen2.5-Omni-7B \citep{xu2025qwen2} thinker model and audio encoder.
The think-aloud module is initialized with Qwen2.5-0.5B-Instruct \citep{qwen2025qwen25technicalreport}.
The unified talker and streaming codec decoder, which are responsible for generating the final audio, are initialized using the CosyVoice 2.0 LLM and the flow matching model \citep{du2024cosyvoice}.

The model is trained for 10,000 steps on 64 H100 GPUs using a batch size of 64.
We employ a learning rate scheduler with a peak value of $1 \times 10^{-5}$, which includes a 500-step warm-up phase followed by an exponential decay.
During model training, we randomly select a turn containing reasoning data from the dialogue to designate as the final turn.
Concurrently, we randomly select either one of the think-aloud responses or the final system response to train the talker.

\subsection{Evaluation}
\label{sec:evaluation}
We evaluate our models on two categories of datasets.
First, to assess reasoning capabilities, we utilize the single-step and multi-step reasoning subsets of the Spoken-MQA benchmark \citep{wei2025towards}.
Correctness is assessed by a GPT-4o judge using a best-of-three majority voting strategy.
Second, to test commonsense knowledge and factuality, we use the Web Questions \citep{berant-etal-2013-semantic}, and TriviaQA \citep{joshi2017triviaqa}.
For this category, we apply the exact match to determine if the ground-truth answer is present in the model's response.
The metric is accuracy for both the Web Questions and TriviaQA test set.

\FloatBarrier
\section{Experimental Results}

\subsection{Main Results}
To evaluate the effectiveness of our approach, we benchmark against two control settings. The \textit{Baseline} represents a model without the think-aloud module, fine-tuned solely on direct-response data, i.e., data that excludes intermediate reasoning steps. The \textit{Baseline with CoT} uses the same backbone but is trained on datasets augmented with reasoning text. This model follows a serial paradigm, generating the full textual reasoning before synthesizing speech, and thus serves as a strong serial reasoning reference, albeit with substantially higher latency. Our proposed model is described as \textit{Baseline with Think-Aloud CoT}.

The two controls serve different purposes. The baseline represents a non-reasoning spoken assistant and is used to measure the gap between our full reasoning-enabled system and a direct-response spoken model.
The baseline with CoT is the more relevant design-level control, since it shares reasoning-augmented supervision with our method but follows a serial think-then-speak pipeline. 

Table~\ref{table:reason} presents the evaluation results on the Spoken-MQA benchmark, which assesses the model's ability to handle both single-hop and multi-hop reasoning in speech.
Compared with the direct-response baseline, our full system achieves substantially higher accuracy on reasoning-intensive tasks, especially on the multi-step subset. This comparison reflects the performance gap between a non-reasoning spoken assistant and our full reasoning-enabled system.
More importantly, when compared with the baseline with reasoning, our method achieves highly comparable performance (87.6\% vs. 88.5\% on average).
This suggests that our framework largely preserves the logical depth and answer accuracy of CoT reasoning, while avoiding the long silent delays of the serial paradigm.

\begin{table}[!htb]
    \caption{Performance on single-step and multi-step reasoning subsets of Spoken-MQA benchmark, measured in accuracy.}
    \label{table:reason}
    \small
    \centering
    \begin{tabular}{lccc}
    \toprule
    \textbf{Models} &  \textbf{Single-Step} & \textbf{Multi-Step} & \textbf{Average} \\
    \midrule
    Whisper-Qwen2.5-7B-Instruct & 81.0 & 68.9 & 75.0 \\
    Whisper-Deepseek-Math-7B-instruct & 85.2 & 78.2 & 81.7 \\
    Whisper-Qwen2.5-Math-7B-Instruct & 88.0 & 86.2 & 87.1 \\
    \midrule
    LLaMA-Omni-7B \citep{fang2024llama} & 29.5 & 10.5 & 20.0 \\
    Qwen2-Audio-7B-Instruct \citep{chu2024qwen2} & 56.2 & 19.2 & 37.7 \\
    Freeze-Omni \citep{wang2024freeze} & 69.0 & 19.8 & 44.4 \\
    GLM-4-Voice \citep{zeng2024glm} & 54.4 & 28.5 & 41.5 \\
    Mini-Omni-Reasoner \citep{xie2025mini} &  85.9 & 60.5 & 73.2 \\
    \midrule
    Baseline & 86.0 & 63.0 & 74.5 \\
    Baseline + CoT & 91.3 & 85.6 & 88.5 \\
    Baseline + \textbf{Think-Aloud CoT (ours)} &91.6 & 83.6 & 87.6 \\
    \bottomrule
    \end{tabular}
\end{table}

We further evaluate general knowledge performance on the Web Questions and TriviaQA datasets under both Speech-to-Speech and Speech-to-Text settings (Table~\ref{table:qa}). Consistent with the findings on Spoken-MQA, our model outperforms the direct-response baseline by a clear margin (e.g., +5.4\% average accuracy in S2S) and remains highly comparable to the baseline with reasoning. Taken together, these results suggest that concurrent think-aloud generation does not materially degrade the factual accuracy of the final response.

\begin{table}[!htb]
    \caption{Speech-to-Speech and Speech-to-Text performance on spoken QA benchmarks, measured in accuracy.}
    \label{table:qa}
    \small
    \centering
    \begin{tabular}{lccc}
    \toprule
    \textbf{Models} & \textbf{Web Questions} & \textbf{TriviaQA}  & \textbf{Average} \\
    \midrule
    \multicolumn{4}{l}{\textit{Speech-to-Speech (S2S)}} \\
    \midrule
    GPT-4o-Realtime \citep{hurst2024gpt} &51.6 & 69.7 & 60.7 \\
    Moshi \citep{defossez2024moshi} & 9.2 & 7.3 & 8.3 \\
    Mini-Omni \citep{xie2024mini} &  12.8 & 6.9 & 9.9 \\
    GLM-4-Voice \citep{zeng2024glm} & 15.9 & 26.5 & 21.2 \\
    LUCY (S2) \citep{gao2025lucy} & 25.6 & 22.9 & 24.3 \\
    Freeze-Omni \citep{wang2024freeze} &26.1 & 25.7 & 25.9 \\
    LLaMA-Omni2-7B \citep{fang2025llama} & 31.3 & - & - \\
    MinMo \citep{chen2025minmo} & 39.9 & 37.5 & 38.7 \\
    MiniCPM-o 2.6 \citep{yao2024minicpm} &40.0 & 40.2 & 40.1 \\
    VITA-Audio \citep{long2025vita} & 41.7 & 42.7 & 42.2 \\
    \midrule
    Baseline & 32.1 & 36.1 & 34.1 \\
    Baseline + CoT & 40.3 & 39.2 & 39.8\\
    Baseline + \textbf{Think-Aloud CoT (ours)}  & 40.3 &  38.7 & 39.5 \\
    \midrule
    \multicolumn{4}{l}{\textit{Speech-to-Text (S2T)}} \\
    \midrule
    Moshi \citep{defossez2024moshi} & 26.6 & 22.8 & 24.7 \\
    LUCY (S2) \citep{gao2025lucy} & 29.3 & 27.0 & 28.2 \\
    GLM-4-Voice \citep{zeng2024glm} & 32.2 & 39.1 & 35.7\\
    LLaMA-Omni2-7B \citep{fang2025llama} & 34.5 & - & - \\
    VITA-Audio  \citep{long2025vita} & 45.0 & 45.9 & 45.5 \\
    \midrule
    Baseline & 33.7 & 39.6 & 36.7\\
    Baseline + CoT & 42.5 & 42.1 & 42.3 \\
    Baseline + \textbf{Think-Aloud CoT (ours)}  & 43.0 & 41.5 & 42.3 \\
    \bottomrule
    \end{tabular}
\end{table}

Tables~\ref{table:reason} and \ref{table:qa} also include results from recent state-of-the-art SLMs for reference. Although direct comparison is difficult due to differences in training data and experimental settings, the results indicate that our method remains competitive relative to contemporary spoken language models.

\FloatBarrier
\subsection{Latency Analyses}
\label{sec:latency}
\begin{table}[!htb]
    \caption{Performance and latency analysis on the Spoken-MQA benchmark. ``DB Stgy.'' is short for dynamic balance strategy.}
    \label{table:latency}
    \centering
    \small
    \begin{tabular}{lcccc}
    \toprule
    \multirow{2}{*}{\textbf{Model / Speed (tok/s)}} & \multicolumn{2}{c}{\textbf{Accuracy (\%)} $\uparrow$} & \multicolumn{2}{c}{\textbf{Latency (s)} $\downarrow$} \\

    \cmidrule(lr){2-3} \cmidrule(lr){4-5}

     & \textbf{Single} & \textbf{Multi} & \textbf{$L_{\mathrm{sil}}$} & \textbf{$L_\mathrm{oh}$} \\
    \midrule
    Baseline & 86.0 & 63.0 & - & - \\
    Baseline + CoT & 91.3 & 85.6 & 12.82 & - \\
    Proposed Model w/o DB Stgy. & 90.7 & 84.5 & 1.03 & 19.92 \\
    \midrule
    \multicolumn{5}{l}{\textit{Proposed Model (varying generation speed)}} \\
    \quad 40 & 91.6 & 83.6 & 0.36 & 4.34 \\
    \quad 80 & 91.6 & 84.2 & 0.05 & 4.16 \\
    \quad 160 & 91.8 & 84.3 & 0.03 & 5.66 \\
    \bottomrule
    \end{tabular}
\end{table}

In real-world deployments, SLMs may run under different hardware and software conditions, leading to varying reasoning speeds. A robust think-aloud mechanism should therefore adapt the duration of generated speech to the speed of internal reasoning.
We evaluate this temporal synchronization using two latency metrics: Unmasked Silence Latency ($L_{\mathrm{sil}}$), the cumulative duration during which reasoning is ongoing but no audio feedback is available to the user, and Articulation Computational Overhead Latency ($L_{\mathrm{oh}}$), the extra delay caused when think-aloud speech continues after the final response becomes ready.

As shown in Table~\ref{table:latency}, the serial reasoning baseline achieves competitive accuracy but suffers from a large ($L_{\mathrm{sil}}$) of 12.82s, making it less suitable for fluid spoken interaction.
In contrast, the ablation without the dynamic balance strategy substantially reduces silence but produces a large ($L_{\mathrm{oh}}$) of 19.92s, indicating that unregulated think-aloud generation can unnecessarily prolong the turn.
Our full model achieves a better trade-off across different reasoning speeds.
Unless otherwise specified, the main latency results are reported under a reasoning generation speed of 40 tok/s.
By dynamically triggering or canceling think-aloud utterances, the proposed strategy suppresses both unmasked silence and articulation computational overhead while maintaining accuracy comparable to the serial reasoning baseline.

\begin{figure}[!htb]
     \centering
     \includegraphics[width=0.9\linewidth]{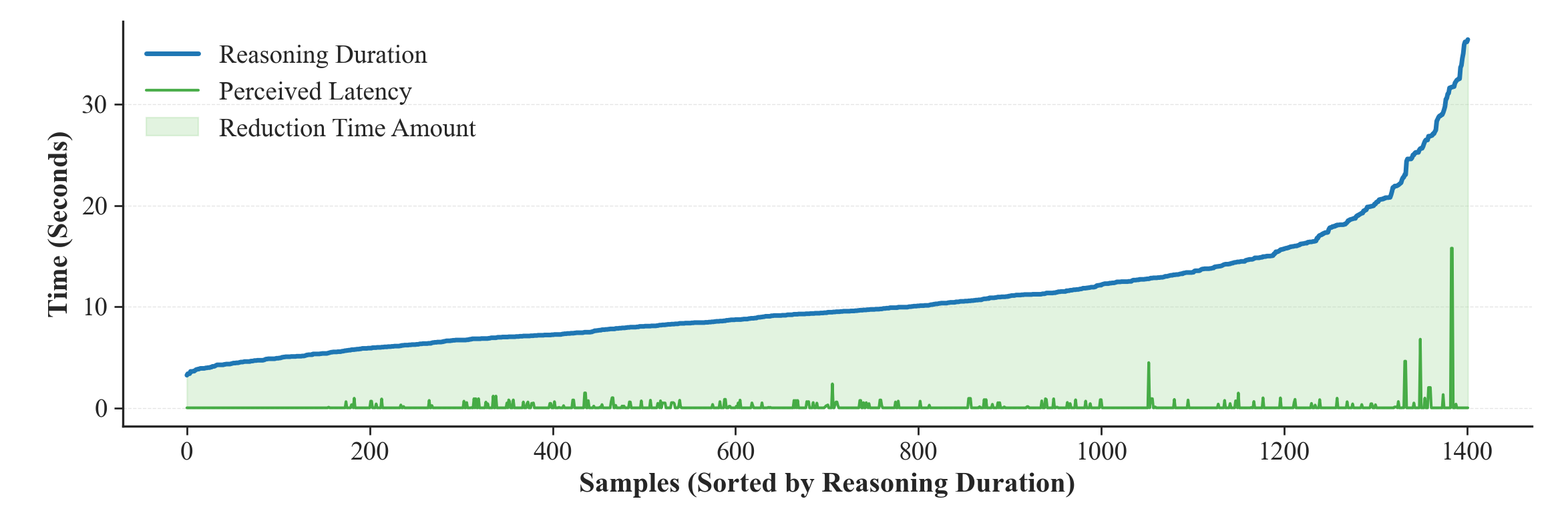}
     \caption{Impact of think-aloud utterances on perceived latency. Samples are sorted by reasoning duration. The blue line shows reasoning duration, the green line shows perceived latency, and the shaded region indicates the silence reduced by think-aloud speech. The results show that the proposed mechanism substantially masks reasoning time and keeps perceived latency low for most samples, even when reasoning takes longer.}
     \label{fig:latency}
\end{figure}

To further analyze the temporal behavior, Figure~\ref{fig:latency} visualizes reasoning duration and perceived latency on the multi-step subset of Spoken-MQA, with samples sorted by reasoning duration. The shaded region indicates the amount of reasoning time masked by think-aloud speech. For most samples, the proposed mechanism keeps perceived latency very low even when the reasoning duration is long, although a small number of outliers still exhibit noticeable wait time.

Figure~\ref{fig:latency} also illustrates why the proposed mechanism is not merely an initial filler.
A single short filler can only mask the beginning of a long reasoning interval.
Repeated generic fillers could further reduce silence, but they would still provide limited information about the task or the model's progress.
Our dynamic triggering mechanism preserves the silence-masking role of fillers while replacing repeated task-independent phrases with progress utterances conditioned on completed reasoning segments.

\FloatBarrier
\subsection{Speech Quality}
\label{sec:speech_qua}
To evaluate speech quality, we compare our model with Qwen2.5-Omni \citep{xu2025qwen2} on the single-step reasoning subset of the Spoken-MQA benchmark \citep{wei2025towards}.
We utilize four metrics: STOI, PESQ, SI-SDR, and NISQA.
The first three metrics are estimated using TorchAudio-Squim \citep{kumar2023torchaudio}.
As shown in Table \ref{table:quality}, higher scores indicate better performance.

Overall, the proposed model achieves intelligibility comparable to that of Qwen2.5-Omni.
However, it provides better perceptual quality, improving PESQ from 3.69 to 3.97 and NISQA from 4.65 to 4.95.
Conversely, the SI-SDR score is slightly lower for the proposed model.
Taken together, these results demonstrate that the speech quality of our model is comparable to that of Qwen2.5-Omni.

\begin{table}[!htb]
    \caption{Performance in terms of speech quality of Qwen2.5-Omni and our proposed model on the single-step reasoning subset of spoken-MQA benchmark.}
    \label{table:speech}
    \label{table:quality}
    \centering
    \small
    \begin{tabular}{lcccc}
    \toprule
    \textbf{Models} & \textbf{STOI} $\uparrow$ & \textbf{PESQ} $\uparrow$ & \textbf{SI-SDR} $\uparrow$ & \textbf{NISQA} $\uparrow$ \\
    \midrule
    Qwen2.5-Omni &  0.99 & 3.69 & 24.71 & 4.65  \\
    Proposed Model & 0.99 & 3.97 & 24.50 & 4.95 \\
    \bottomrule
    \end{tabular}
\end{table}

\FloatBarrier
\subsection{Human Evaluation}
To complement the objective evaluations, we conduct a human evaluation on 50 randomly sampled test cases from the multi-step subset of Spoken-MQA benchmark, with 3 annotators per case.
For each sample, annotators compare the output from our proposed model with the output that remains silent during thinking and only outputs the final response.
In a blind A/B setting, they are asked to choose (1) which system they prefer overall and (2) which system feels more responsive.
In addition, they rate the fluency/naturalness of the proposed model output on a 5-point Likert scale (1: very unnatural, 3: Acceptable, 5: very natural).

The proposed model is preferred overall in 85.3\% of the cases and is judged to be more responsive in 99.3\% of the cases, while achieving a fluency/naturalness score of 4.01.
Qualitatively, we observe that the inserted think-aloud responses are generally consistent with the final responses, and the transition to the final answer is typically smooth rather than abruptly cut off.
Representative qualitative examples are provided in Section~\ref{sec:case_study}.

\subsection{Qualitative Examples}
\label{sec:case_study}
We present a standard operating example and a separate illustrative future-use scenario.

\subsubsection{Standard Think-Aloud Operation}

\begin{figure}[!htb]
    \centering
    \includegraphics[width=0.9\linewidth]{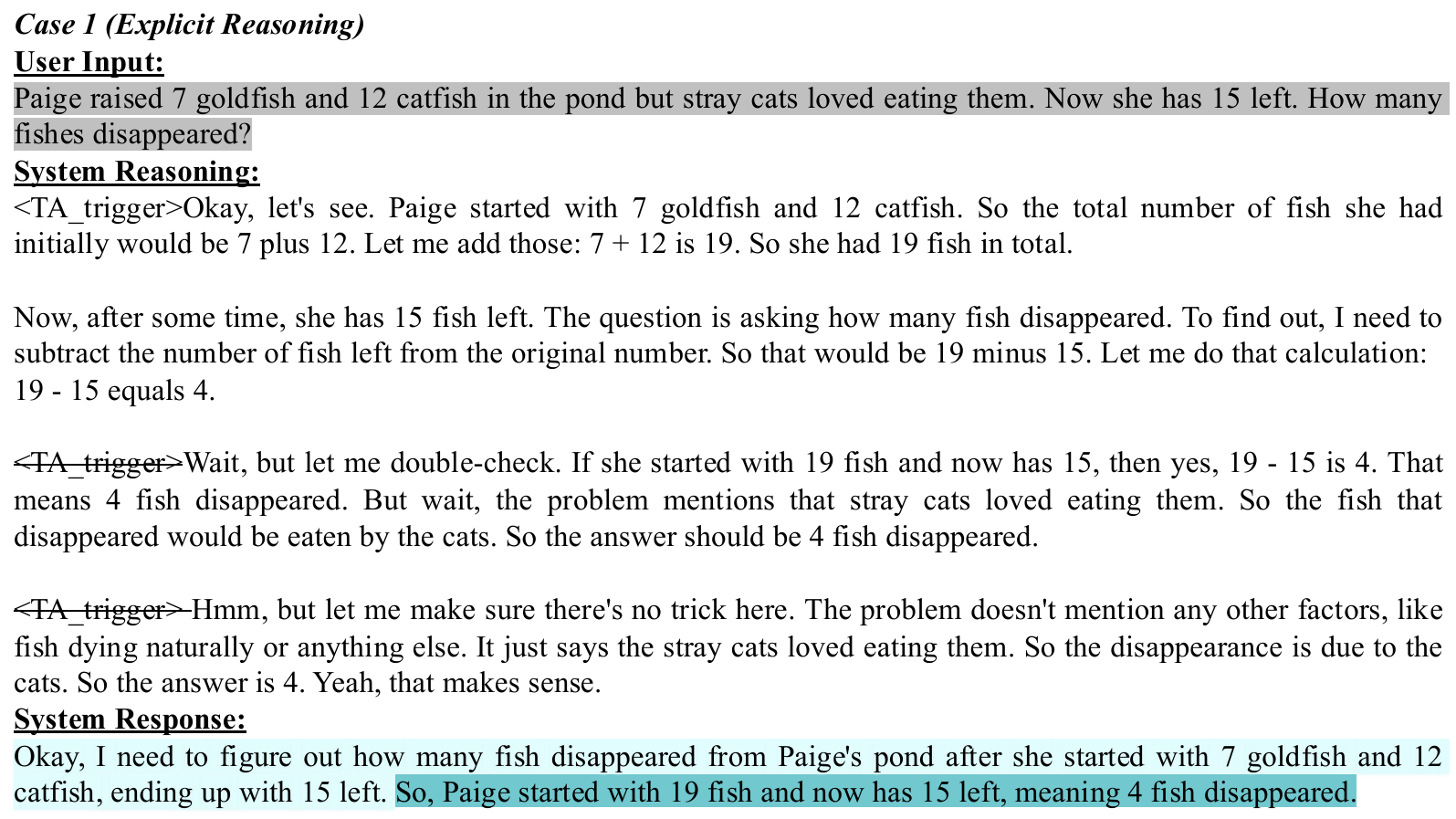}
    \caption{Case 1 demonstrates thinking while speaking.
\hlbox{usergray}{Gray} backgrounds represent User Input.
\hlbox{tablue}{Light Blue} denotes the think-aloud responses, while \hlbox{finalblue}{Dark Blue} highlights the final response derived from the reasoning process.
The \stoken{TA\_trigger} with a strike-through indicates that the trigger is canceled by the dynamic balance strategy.}
    \label{fig:case1}
\end{figure}

Figure~\ref{fig:case1} illustrates the standard operating mode of the proposed model.
During the thinker reasoning process, a total of three \stoken{TA\_trigger}s are generated.
At the current reasoning speed, the first utterance is sufficient to cover the reasoning period, so the second and third triggers are canceled by the dynamic balance strategy.
Instead of remaining silent for a long period and then producing a direct answer, the system first generates a think-aloud response, reducing the silent gap between the query and the final solution before delivering the final response.

\Needspace{12\baselineskip}
\subsubsection{Simulated Interruption for Error Correction}

\begin{figure}[!htb]
    \centering
    \includegraphics[width=0.9\linewidth]{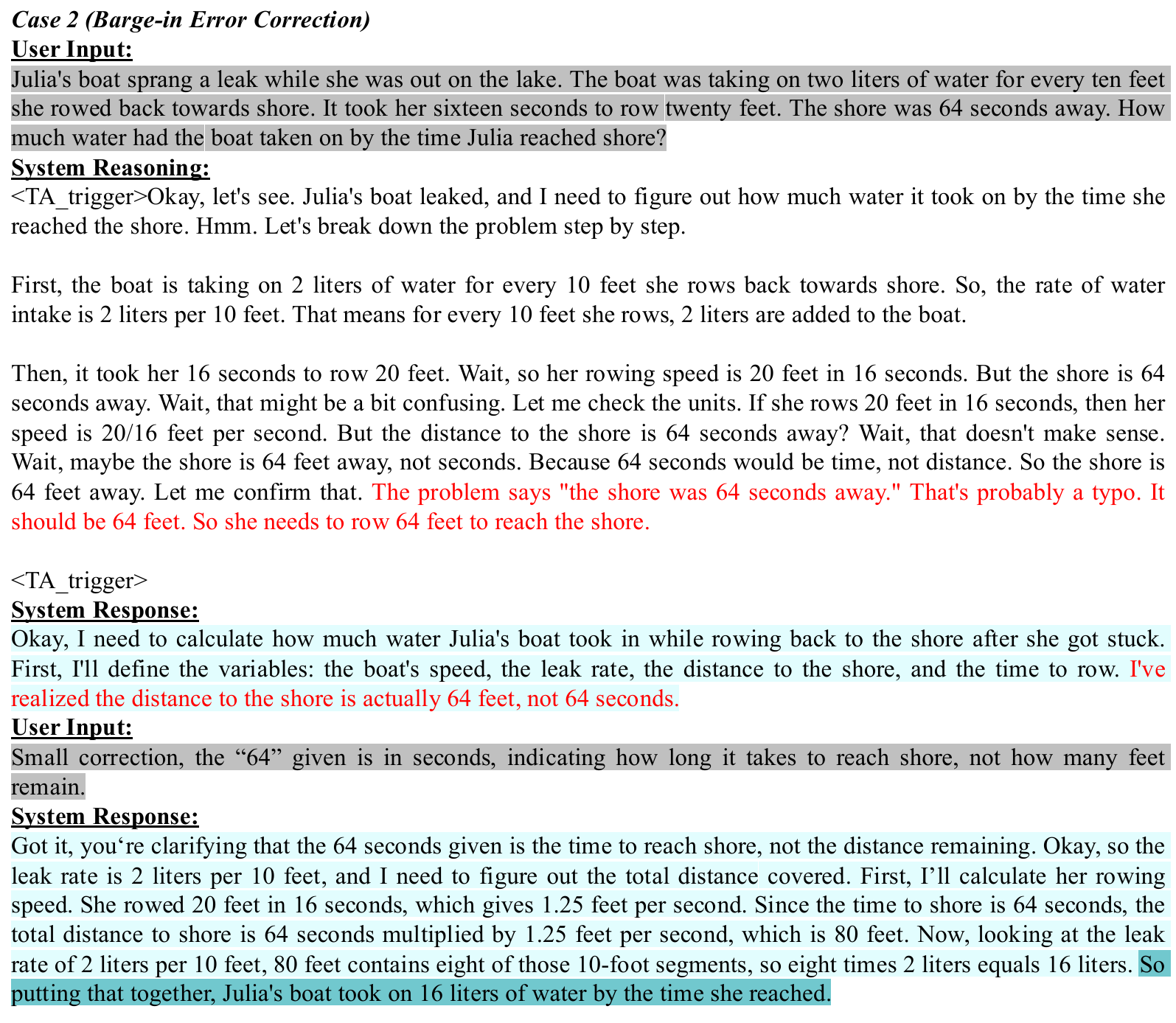}
    \caption{Case 2 illustrates a simulated interruption scenario for error correction.
\textit{This example is intended to show a possible deployment setting in which exposed think-aloud speech may allow earlier user intervention when an error is detected in the reasoning stream.} The scenario assumes an external interruption mechanism (e.g., VAD or turn-taking control), which is not implemented or evaluated in this work. For readability, we omit the reasoning text of the second turn.}
    \label{fig:case2}
\end{figure}

Figure~\ref{fig:case2} illustrates a possible deployment setting in which exposed think-aloud speech may allow earlier user intervention when an error is detected in the reasoning stream.
It is included as an illustrative future-use scenario rather than as a validated capability of the current system.
The scenario assumes an external interruption mechanism, such as VAD or turn-taking control, which is outside the scope of this work and is not implemented or quantitatively evaluated here.

\section{Related Work}
\subsection{Spoken Language Models}
Existing SLMs typically generate speech through either interleaved text-speech decoding \citep{zeng2024glm,li2025baichuan} or the Thinker-Talker architecture \citep{xu2025qwen2,ding2025kimi,fang2025llama}.
In the interleaved paradigm, the backbone alternately generates text and speech chunks, where text serves as a transcription that guides subsequent audio.
In the Thinker-Talker paradigm, a thinker first produces text and intermediate representations, which a talker then converts into speech tokens. Despite these differences, both paradigms tightly couple generated text with speech output, effectively treating text as a script for synthesis.
Our work instead equips the SLM with intrinsic reasoning capabilities and introduces a dedicated think-aloud module that verbalizes selected reasoning-grounded milestones as natural progress utterances during inference.

A separate line of work targets full-duplex spoken interaction, where the model can listen and speak simultaneously.
Systems such as Moshi~\citep{defossez2024moshi} use parallel audio streams to support barge-in and overlapping speech.
These approaches improve responsiveness along the listen-while-speak dimension, whereas our framework targets a different source of latency: the internal reasoning delay after a user query has been received.
We therefore study the complementary think-while-speak dimension, where a primary reasoning stream continues in the background while a speaking stream provides audible progress feedback.
Our current system does not address barge-in directly, but its decoupled reasoning-and-speaking design could be integrated with full-duplex audio interfaces in future work.

\subsection{Reasoning in Spoken Language Models}
Integrating CoT reasoning into SLMs remains challenging, as it requires balancing reasoning capability, modality alignment, and low-latency interaction.
Early efforts \citep{xie2025audio,wen2025sari,li2025reinforcement} primarily study reasoning over audio or speech inputs in text-centric settings, while the problem of maintaining fluid spoken interaction during long reasoning remains less explored.
To mitigate this issue, recent approaches such as STITCH and Mini-Omni-Reasoner \citep{chiang2025stitch,xie2025mini} reduce latency by interleaving reasoning and spoken-response generation during inference, allowing intermediate reasoning to be incorporated while speech is produced incrementally.
These methods typically realize thinking and speaking within a predefined interleaved generation schedule, such as chunk-level alternation or token-level interleaving.

Related parallel-processing paradigms have also been explored.
For example, \citet{Shih2025CanSL} propose a Thinking-while-Listening framework that reduces response delay by initiating text-based reasoning before the user has finished speaking.
AsyncVoice Agent \citep{lin2025asyncvoice} similarly explores asynchronous voice interaction with live reasoning streams, but focuses on an external reasoning backend and an interruptible explanation interface rather than an integrated spoken language model.

In a related but distinct direction, our framework focuses on thinking while speaking: it adopts a two-stream asynchronous design in which a reasoning thinker continuously advances the main reasoning trajectory, while a separate think-aloud module verbalizes selected reasoning milestones for a unified talker to synthesize.
By explicitly decoupling reasoning progression from speech delivery, our approach enables the two streams to proceed concurrently, rather than following a predefined chunk-level or token-level interleaving schedule.

This design should not be interpreted as universally superior to interleaving-based approaches.
Instead, it represents a complementary architectural trade-off.
Fixed chunk-level or token-level interleaving provides a direct and effective mechanism for incremental generation, whereas our decoupled design aims to offer more flexible runtime scheduling when reasoning duration and speech duration are mismatched.
Crucially, guided by a dynamic balance strategy, our system can adapt to varying inference speeds by triggering additional progress speech under audio starvation or canceling pending utterances when the final response becomes ready, without relying on a fixed generation ratio.
As a result, the proposed Think-Aloud mechanism provides selected reasoning-grounded progress feedback as speech, reducing prolonged silent gaps and improving the responsiveness of spoken interaction across diverse deployment conditions.
A direct empirical comparison between asynchronous two-stream decoupling and interleaving-based spoken reasoning remains an important direction for future work.

\section{Conclusion}

In this work, we presented an asynchronous think-aloud framework to address the tension between reasoning latency and responsiveness in spoken interaction.
By coordinating a primary reasoning stream with a lightweight think-aloud stream, our framework provides selected reasoning-grounded progress feedback while internal reasoning continues in the background.
Experiments show that the proposed method substantially reduces unmasked silence and improves turn responsiveness, while preserving most of the reasoning benefits of a serial think-then-speak system.
Overall, our method provides a practical design framework for building more responsive spoken reasoning agents.
Future work will study controlled comparisons with generic fillers, fully interactive user evaluation, and adaptive policies that shorten or skip reasoning when unnecessary.

\section{Acknowledgment}
We thank Yingru Liu for helpful discussions and feedback on this project.

\bibliographystyle{assets/plainnat}
\bibliography{paper}

\clearpage
\appendix
\section{Data-Construction Prompts}
\label{sec:prompts}
This appendix provides the prompts used to prepare reasoning data and think-aloud responses following the pipeline in Section~\ref{data_prep}.

\subsection{Reasoning Data Generation}
\label{sec:reasoning_prompts}

The prompts for the three construction steps are provided below.

\subsubsection{Turn Selection}
\begin{Verbatim}[fontsize=\footnotesize,breaklines=true,breakanywhere=true,frame=single]
You are a helpful assistant whose job is to classify user inputs according to their complexity for a speech-based LLM. You will be given two pieces of information:
1. **Conversation History**: A list of alternating user and assistant turns leading up to the current input. None means no conversation history.
2. **Current User Input**: A single new user query.

You must determine whether the **Current User Input** is "Complex," meaning **it meets at least one** of the following criteria:
- **Human-Thinking Complexity**: The query requires non-trivial reasoning or thought, even for a human (e.g., multi-step planning, abstract reasoning, or detailed technical explanations).
- **Response-Length Complexity**: The natural answer would be so long or detailed that it is unwieldy for a real-time spoken dialogue (e.g., multi-paragraph exposition, lengthy code walkthroughs, extensive comparisons).

**Output Format:**
Return exactly one JSON object with two fields:
- "complex": true|false,
- "reasons": [list of strings]

complex is true if **any** of the two criteria apply, otherwise false. reasons lists which criteria were triggered; valid values are "human_thinking" and "lengthy_response".

Example:

[Conversation History]

User: "How do I bake a chocolate cake?"

Assistant: "Sure-follow these steps: ..."

[Current User Input]

"Can you explain the Maillard reaction in baking and also compare it to caramelization?"

Your output should be JSON form with two keys:
- "complex": true,
- "reasons": ["human_thinking", "lengthy_response"]

Now evaluate this input:

[Conversation History]

{dialog_history}

[Current User Input]

{user_input}

Please produce only the JSON result. Do not include any other text.
\end{Verbatim}

\Needspace{14\baselineskip}
\subsubsection{Reasoning Generation}
\begin{Verbatim}[fontsize=\footnotesize,breaklines=true,breakanywhere=true,frame=single]
[Dialogue History]

{dialog_history}

[Input]

{user_input}
\end{Verbatim}

\subsubsection{Validation and Filtering}
\begin{Verbatim}[fontsize=\footnotesize,breaklines=true,breakanywhere=true,frame=single]
You are given the following:
- **Dialogue History**: Previous exchanges between the user and assistant (may be empty).
- **User Question**: The new user question.
- **Ground Truth Answer**: The correct answer.
- **Reasoning Process**: A multi-step explanation generated by another model, which may include intermediate mistakes that are later identified and corrected.

**Task**:
Determine whether the reasoning process is logically consistent overall and sufficient to justify the ground truth answer, **even if intermediate steps contain errors that are later corrected**. Use any relevant information from the dialogue history.
- The final answer must be fully justified by the reasoning process as it concludes.
- Ignore minor surface errors (grammar, style, length) unless they affect logic.
- If the reasoning process identifies and corrects its own mistakes, and the final logic is sound and sufficient for the ground truth answer, select "Yes."
- If there remain uncorrected mistakes, unjustified steps, or missing information that prevent reaching the ground truth answer, select "No."
- Output strictly in the JSON format (no extra text) with two fields:
- "Answer": "Yes" or "No",
- "Explanation": "A brief explanation of your reasoning (1-3 sentences)."

**Input**:
[Dialogue History]

{dialog_history}

[User Question]

{user_input}

[Ground Truth Answer]

{assistant_output}

[Reasoning Process]

{reasoning}
\end{Verbatim}

\subsection{Think-Aloud Response Generation}
\label{sec:ta_prompts}

The following prompts implement trigger identification, utterance generation, and final-response rewriting.

\subsubsection{Trigger Identification and Filtering}
\begin{Verbatim}[fontsize=\footnotesize,breaklines=true,breakanywhere=true,frame=single]
**You are an expert in analyzing chain-of-thought (CoT) reasoning for conversational AI.**

I'll give you three pieces of information:

1. **Dialogue History**
   A chronological list of past user and assistant turns.
2. **Current User Input**
   The user's most recent message.
3. **CoT Reasoning**
   The assistant's in-progress chain of thought, split into numbered paragraphs.

Your job is to examine each numbered paragraph of reasoning and decide whether, **at its end**, the assistant has enough insight to utter a "think-aloud" response.

Output **only** valid JSON in this exact structure:

```json
[
  {
    "paragraph": <integer>,    // the paragraph number where speaking is now possible
    "reason":  "<string>"      // one-sentence justification
  },
]
```

**Rules:**

* Trigger speech only when a coherent sub-inference or decision is complete.
* Do **not** include any prose outside the JSON array.
* Each object must have exactly two fields:

  1. "paragraph" (integer)
  2. "reason" (string)

**Example**

*Input:*

[Dialogue History]

None

[Current User Input]

What's the best place to find international workers to build a profile for me?

[CoT Reasoning]

**1.** Okay, the user wants to know the best place to find international workers to build a profile. Let me break this down. First, I need to clarify what they mean by "build a profile." It could be a professional profile like LinkedIn, a personal portfolio, or maybe something else. Since they mentioned "international workers," they're probably looking for freelancers or agencies from different countries.

**2.** I should consider platforms where you can hire freelancers globally. Upwork and Fiverr come to mind immediately. They have a wide range of professionals from various countries. Maybe also Toptal for higher-end talent. But maybe the user is looking for something specific, like a resume writer or a personal branding expert. It's worth mentioning niche platforms too, like Behance for designers or GitHub for developers if it's a tech profile.

**3.** They might not be aware of LinkedIn ProFinder or other specialized services. Also, language skills might be a factor, so platforms that allow filtering by language or region could be useful. I should also think about whether they want individual freelancers or an agency. Some platforms cater more to agencies, like Crew or Guru.

**4.** Another angle: maybe they need help with localization, so mentioning platforms that have professionals skilled in different regions could help. Also, payment methods and currency considerations might be important when hiring internationally. Safety and vetting processes on these platforms are important too---ensuring the workers are reliable.

**5.** I should list the top platforms, explain their strengths, and maybe give a brief tip on what to look for when hiring internationally. Don't forget to mention checking reviews and portfolios. Maybe add a note about communication tools and time zones. Alright, that covers the main points. Time to structure this into a clear, concise answer with options and considerations.

*Expected Output:*

```json
[
    {
        "paragraph": 2,
        "reason": "At this point, the assistant has identified key freelancing platforms like Upwork and Fiverr for finding international workers."
    },
    {
        "paragraph": 3,
        "reason": "The assistant has evaluated specialized services and hiring considerations such as freelancer versus agency options."
    },
    {
        "paragraph": 4,
        "reason": "The assistant has covered international factors like localization and safety, forming a coherent sub-inference on hiring challenges."
    },
    {
        "paragraph": 5,
        "reason": "The assistant has synthesized all insights and can now deliver a structured response with recommendations."
    }
]
```

**Inputs:**

[Dialogue History]

{dialog_history}

[Current User Input]

{user_input}

[CoT Reasoning]

{reasoning}
\end{Verbatim}

\subsubsection{Think-Aloud Response Generation}
\begin{Verbatim}[fontsize=\footnotesize,breaklines=true,breakanywhere=true,frame=single]
You are an expert spoken-dialogue-system researcher.
Your task is to generate a **think-aloud** utterance **speaking to the user**, representing the system's internal reasoning **up to and including Paragraph <N>** of its chain-of-thought (CoT).

#### Inputs
- DIALOGUE_HISTORY: {dialog_history}
- USER_INPUT: {user_input}
- COT_REASONING_PARAGRAPHS: {reasoning}
- TARGET_PARAGRAPH (N): {target_para}
- FINAL_SYSTEM_RESPONSE: {assistant_output}

#### Requirements of the output
1. **Scope** -Reference only information found in CoT paragraphs 1 through N (do *not* anticipate paragraph N+1).
2. **Tone** - Sound like a partially formed spoken thought: natural, informal.
3. **Coherence** - Stay logically and content-wise consistent with the FINAL_SYSTEM_RESPONSE (no contradictions).
4. **Format** - Plain text, **only one long sentence** with **no more than 10 words**, no lists/JSON/markdown, no explicit paragraph numbers.

Begin now.
\end{Verbatim}

\Needspace{42\baselineskip}
\subsubsection{System Response Rewriting}
\begin{Verbatim}[fontsize=\footnotesize,breaklines=true,breakanywhere=true,frame=single]
You are an expert in conversational AI and dialogue flow. Your task is to revise a system's final response to make it more coherent and natural, seamlessly continuing from its preceding "think-aloud" monologue.

**Context:**
The AI system first verbalizes its reasoning process (the "Think-aloud Response") and then delivers a final, conclusive answer (the "Original System Response"). There is currently a coherence gap between these two parts.

**Your Goal:**
Rewrite the `[Original System Response]` to create an `[Improved System Response]`.

**Instructions and Constraints:**
1.  **Ensure Coherence:** The `[Improved System Response]` must be a logical and smooth continuation of the `[Think-aloud Response]`. It should feel like the natural conclusion to the thoughts that were just spoken.
2.  **Preserve Core Meaning:** The essential information and intent of the `[Original System Response]` must be fully preserved. Do not add new factual information or contradict the original answer.
3.  **Conversational Tone:** The output should be natural and suitable for a spoken dialogue system. Avoid robotic or overly formal language.
4.  **Conciseness:** Be clear and to the point, just as a human would conclude their thoughts.
5.  **Strict Output Format:** You MUST output ONLY the text of the revised system response. Do not include any extra text, explanations, acknowledgements (like "Sure, here it is:"), or labels (like "[Improved System Response]:"). Your entire output will be the response itself.

**Input:**

**[User Question]:**

{user_question}

**[Think-aloud Response]:**

{think_aloud_response}

**[Original System Response]:**

{original_system_response}

**Output:**

**[Improved System Response]:**
\end{Verbatim}

\end{document}